\documentclass[a4paper]{article}
\usepackage{iwslt18,amssymb,amsmath,epsfig}
\def\reg{{\rm\ooalign{\hfil
     \raise.07ex\hbox{\scriptsize R}\hfil\crcr\mathhexbox20D}}}
     
\usepackage{times}
\usepackage{url}
\usepackage{latexsym}
\usepackage{graphics}
\usepackage{graphicx}
\usepackage{xcolor}
\usepackage{enumitem}
\setitemize{noitemsep,topsep=0pt,parsep=0pt,partopsep=0pt}

\usepackage{todonotes}

\newcommand{\petpw}{PETpW}
\newcommand{\petpwh}{PETpW}
\newcommand{\mted}{MT'ed}%
\newcommand{\peed}{PE'ed}

\title{Estimating post-editing effort: a study on human judgements, task-based and reference-based metrics of MT quality}
\name{\emph{Carolina Scarton}$^1$, \emph{Mikel L.\ Forcada}$^2$, \emph{Miquel Espl\`{a}-Gomis}$^2$, \emph{Lucia Specia}$^{1,3}$}
\address{$^1$ Department of Computer Science, University of Sheffield, Sheffield S1 4DP, UK \\
$^2$ Dept.\ Llenguatges i Sist.\ Inform., Universitat d'Alacant, 03690 St.\ Vicent del Raspeig, Spain \\
$^3$ Department of Computing, Imperial College London, London SW7 2AZ, UK. \\
\\
{\small \tt c.scarton@sheffield.ac.uk, mlf@ua.es, mespla@dlsi.ua.es, l.specia@imperial.ac.uk }
}
\begin{document}
\maketitle
%
%\blfootnote{CS and MLF are both first authors with equal contribution.}
\begin{abstract}

Devising metrics to assess translation quality has always been at the core of machine translation (MT) research. 
Traditional automatic reference-based metrics, such as BLEU, have shown correlations with human judgements of adequacy and fluency and have been paramount for the advancement of MT system development. 
Crowd-sourcing has popularised and enabled the scalability of metrics based on human judgments, such as subjective \emph{direct assessments} (DA) of adequacy, that are believed to be more reliable than reference-based automatic metrics. 
Finally, task-based measurements, such as post-editing time, are expected to provide a more detailed evaluation of the usefulness of translations for a specific task. 
Therefore, while DA averages adequacy \emph{judgements} to obtain an appraisal of (perceived) quality independently of the task, and reference-based automatic metrics try to objectively estimate quality also in a task-independent way, task-based metrics are \emph{measurements} obtained either during or after performing a specific task. 
In this paper we argue that, although expensive, task-based measurements are the most reliable when estimating MT quality in a specific task; in our case, this task is post-editing. To that end, we
report experiments on a dataset with newly-collected post-editing indicators
 and show their usefulness when estimating post-editing effort. Our results show that task-based metrics comparing machine-translated and post-edited versions are the best at tracking post-editing effort, as expected. These metrics are followed by DA, and then by metrics comparing the machine-translated version and independent references. 
 We suggest that MT practitioners should be aware of these differences and acknowledge their implications when deciding how to evaluate MT for post-editing purposes.

\end{abstract}
\section{Introduction}

%\section{Introduction}\todomlf{subsections?}\todocs{I am not sure about the word `predict'. Are we predicting anything? Maybe replace `to predict' by `as a proxy of' / `to estimate'}\todomlf{Yes, chase and change}

Assessing the quality of the output of machine translation (MT) systems has been a widely explored topic in the last two decades. As with other applications outputting language (e.g.\  text summarisation), quality assessment of MT is challenging and highly dependent on the purpose of the translation. Therefore, the quality of machine translated (\mted{}) texts may 
%be perceived differently depending 
depend
on 
their usage.
%the usage of such text. 
Table \ref{tab:example-assess} shows two 
machine translations into English, and their respective post-edited versions. In Example~1, the \mted{} version has a different meaning from that of the original sentence: readers may be led to believe that \textit{the product} is good and they should buy it, whilst the correct recommendation is against buying the product. Although this sentence would be problematic for an end user, it is rather simple to correct by a post-editor (only one word needs to be added). Example~2, on the other hand, shows a sentence where an end user can understand the \mted{} version with little effort, even though it contains multiple errors. A post-editor, however, would need to perform at least five word-level edit operations in order to transform the machine translation into the post-edited version.

%Table \ref{tab:example-assess} shows two original sentences, their machine translations into English, and their respective post-edited versions. In Example 1, the machine translated version has a different meaning from that of the original sentence: the reader may be led to believe that \textit{Oramas} is the \textit{monarch} and that \textit{he} expressed \textit{his convictions} about a topic. However, the correct meaning (expressed by the post-edited version) is that \textit{Oramas} is a different person that expressed \textit{her opinion} to the \textit{monarch}.\footnote{\emph{Oramas} refers to  \emph{Ana Mar\'ia Oramas}, a female member of the Spanish parliament, and, therefore, the pronoun \textit{his} was also corrected to \textit{her} in the post-edited version.} Although this sentence would be problematic for an end user, it is considerably simple to correct by a post-editor (only three words needed to be added/replaced). 

\begin{table*}[ht]
    \centering
    %\scalebox{0.8}{
    \scalebox{0.9}{
    \begin{tabular}{l|p{7cm}|p{7cm}}
            & Example 1 & Example 2\\
        \hline
        Reference & Do \textcolor{red}{\textbf{not}} buy this product, it's their craziest invention! & The \textcolor{red}{\textbf{battery}} \textcolor{blue}{\textbf{lasts}} \textcolor{magenta}{\textbf{6 hours}} and it can be \textcolor{olive}{\textbf{fully recharged}} in \textcolor{orange}{\textbf{30 minutes}}.\\
         %SRC & \textit{Oramas transmiti\'o al Monarca su convicci\'on de que ``pronto habr\'a un proceso electoral'', porque ve poco probable que Rajoy o S\'anchez puedan formar Gobierno.} & \textit{Das ist zwar hart, aber ich hoffe, dass ich den F\"uhrerschein bis zu meinem 28. Geburtstag in 2 Monaten geschafft habe.} \\
         \hline
         MT & Do buy this product, it's their craziest invention! & \textcolor{magenta}{\textbf{Six-hours}} \textcolor{red}{\textbf{battery}}, \textcolor{orange}{\textbf{30 minutes}} to \textcolor{olive}{\textbf{full charge}} \textcolor{blue}{\textbf{last}}.\\
         %MT & \textit{\textbf{\textcolor{red}{Monarch}} Oramas transmitted to \textbf{\textcolor{red}{his}} conviction that ``soon there will be an election'' because looks unlikely that Rajoy or Sanchez can form a government.} & \textit{Although \textbf{\textcolor{red}{this}} is tough, \textbf{\textcolor{red}{but}} I hope I driver's license up to my 28th birthday\textbf{\textcolor{red}{'ve done}} in 2 months.} \\
         \hline
         %PE & \textit{Oramas transmitted to \textbf{\textcolor{blue}{the monarch her}} conviction that ``soon there will be an election'' because it looks unlikely that Rajoy or Sanchez can form a government.} & \textit{Although \textbf{\textcolor{blue}{it}} is tough, I hope I \textbf{\textcolor{blue}{will have my}} driver's license up to my 28th birthday\textbf{\textcolor{blue}{,}} in 2 months.} \\
    \end{tabular}}
    \caption{Examples of \mted{} sentences and their \peed{} versions}
    \label{tab:example-assess}
\end{table*}

These examples illustrate how sensitive MT evaluation is to purpose. Nirenburg \cite{Nirenburg:1993} argues that MT can be classified into two groups according to its purpose: dissemination or assimilation. MT for dissemination is expected to be either 
%perfect 
ready as is
or adequate for post-editing, since the purpose is publication. 
%Differently, 
In contrast,
MT for assimilation has the purpose of communication: the \mted{} text does not need to be 
%perfect 
grammatically correct
as long as the reader can understand its message. In this paper we will focus on MT for dissemination; more specifically, 
%in the case where 
when 
MT is used for post-editing (PE). 
%\todomlf{The above paragraph has been reworded to avoid the word \emph{perfect}.}
%\todomeg{Check this new paragraph about the human-parity paper}
% Removed as this is beside the point I think
PE is the task of 
%correcting 
editing
%
%machine translated 
\mted{} texts, a
% texts to make them adequate for publication. Nowadays this is 
%a 
common practice among translation providers, where the aim is to improve productivity and, consequently, reduce translation costs. However, %previous work has identified that 
when \mted{} sentences contain too many problems, it may be easier to translate from scratch than to post-edit MT (this is indeed often reported by translators \cite{sanchez2016machine}). 

Some recent work has claimed that state-of-the-art MT can be used for dissemination without supervision, with a well-known example claiming to have achieved \emph{human parity} for Chinese-to-English translation \cite{hassan2018}. However, in-depth revisions of this work~\cite{laubli2018,toral-etal:2018} suggest that we are still far from achieving human performance and, therefore, PE will still remain a key task in the translation industry. Therefore, finding ways of estimating the quality of \mted{} texts in terms of PE effort is a highly desirable feature (it would, for example, allow accurate budgeting of a translation job) and it is also a relevant topic for research in MT.

%also \cite{huang2014improving}). %\todomlf{I need to look for a reference here \textbf{MEG:} What about \cite{sanchez2016machine}: they don't say so explicitly but show that there is a clear relation between MT quality and PE time}

According to Krings \cite{Krings2001}, PE effort has three dimensions: temporal, cognitive and technical. The temporal dimension is the one most easily related to professional productivity or throughput:
one just has to directly measure the time spent by the post-editor in transforming the MT output into 
%a good quality 
an adequate
\peed{} version. PE time for a sentence may be expected to increase roughly linearly with the total number of words in a sentence; therefore, PE time is normalised by dividing it by the number of words in the \mted{} sentence. The measured ratio between PE time and the number of words  in the segment (\petpw{}) can be directly used to assess the effort of post-editing a segment. The main drawback of extracting PE time is that it is relatively more expensive and requires post-editors to avoid breaks during the editing of a given sentence.

Previous work has proposed several ways to address this issue. For example, the shared task 
%in 
on
quality estimation (QE) of MT organised yearly as part of WMT conferences started with the purpose of training models to predict perceived PE effort, moving later to predicting more accurate measurements such as actual PE time and the \emph{translation edit rate} (TER) \cite{snover2008terp}
%\todocs{is this the best reference to TER?}%\todomlf{This is the one cited by \cite{Snover-etal:2006}; can probably be removed.} 
observed when comparing a \mted{} sentence and its post-edited (\peed{}) counterparts, called \emph{human-targeted TER} or HTER \cite{Snover-etal:2006}. HTER 
%indirectly measures 
gives an indirect indication of 
the effort needed to transform a \mted{} sentence into its \peed{} version.
%\todomlf{We needed to define TER and then HTER, hence the rewording}

%OLD TEXT
%\emph{human translation edit rate} (HTER) \cite{Snover-etal:2006}. HTER works as the %TER metric, however, the comparisons are made between post-edited sentences and %machine translation. Therefore, the idea is to indirectly measure the effort into %transforming a machine translated sentence into its post-edited counterpart.

Despite its popularity, HTER has %particularly 
been subject to criticism: 
%there are several criticism to HTER. 
%Mainly the work of \cite{Graham-etal:2016}
Graham et al.~\cite{Graham-etal:2016}
%heavily 
 criticise this metric and contend that subjective direct assessments (DA) of adequacy 
 %(see section~\ref{sec:pevsda}) 
 %\footnote{The degree to which ``the machine-translated segment expresses the meaning of the reference segment in the target language'' \cite{Graham-etal:2016}.}% 
 are more reliable than HTER measurements \cite{Graham-etal:2014,Graham-etal:2016-da}. 
 %moved from deleted section
 %\cite{Graham-etal:2016-da} 
 They
 define adequacy as the degree to which the \mted{} segment expresses the meaning of the reference segment in the target language. Adequacy is therefore assessed in the target language, monolingually. Their DA is a combination of %multiple
 many 
 independent human judgements of adequacy for a given sentence (in a 0\%--100\% scale) into a single score ---standardised to zero mean and unit standard deviation after low-quality assessments are filtered out.
 %It is, however, not flawless. First, in 
 %In
 %order to make it less prone to annotators' biases, a large number of annotators is required.
 %needed, which may not always be feasible.
%Second, annotators may be biased towards the reference translation being shown, even though other possible translations may also be adequate \cite{fomicheva2016reference}, or the provided reference can be incorrect \cite{toral-etal:2018}. %\todols{check if one of human-parity critique papers (EMNL18, WMT8) mention that explicitly and cite it: some references are wrong}.
%Finally, it is not clear whether or not adequacy scores are reliable for predicting MT usefulness in a production scenario. For example, what does a DA score of 50\% mean for a translator who has to decide whether post-editing a translation is better than translating it from scratch?
 %Although we
 %We agree that DA may be a useful and reasonably cheap way of assessing subjectively the adequacy of MT output;
 %it is not clear that DA and HTER (or other metrics based on \peed{} data) measure the same aspect of quality and Graham et al. \cite{Graham-etal:2016} do not answer this question. 
 %in this paper we set out to study how good DA, HTER, and other metrics are at approximating PE effort, 
 
 %In addition, as explained above, 
 As discussed above, MT should be evaluated according to its purpose. Nevertheless, previous work has disregarded this assumption by using DA as gold standard for tasks where PE effort is the aspect of quality to be assessed \cite{Graham-etal:2016,Graham-etal:2017,Bentivogli-etal:2018}. Although we agree that DA may be a useful and reasonably cheap way of assessing subjectively the adequacy of MT output, in this paper we provide an in-depth analysis of ways to assess PE effort, with a focus on reducing PE time, a highly desirable feature by the translation industry.
%
%\todomlf{We should (a) avoid attacking DA directly and clearly recognizing its value as an average of subjective attempts to predict the usefulness of MT regardless of task; (b) try to compare the cost of DA and the cost for HTER to make for a stronger point.}
%\todocs{(a) I don't think DA is being attacked: I tried to say that DA measures something and HTER measures something else. (b) DA is much cheaper than HTER or any other metrics of PE effort. We can include this as an advantage to DA if you want}
%

We propose to assess the usefulness of metrics according to their ability to \emph{rank} translations based on the time that would be required to post-edit them. 
This has a very practical application in the translation industry, 
%since a post-editor or a company may be interested in estimating the time (or effort) a PE job would take\todols{why would ranking give you that? if you need to post-edit the entire batch, ranking won't help}. In addition, 
where knowing which segments are easier to post-edit and which are the most difficult would allow a project manager to select 
%annotators 
post-editors accordingly, perhaps sending segments estimated as ``easier'' to less experienced (or cheaper) translators and/or sending the ``most difficult'' segments to experienced translators 
or
to be translated from scratch.
%\todomlf{There is a bit of repetition with material with the first paragraph of section 4}
%%%%%%%%%%%%%%%%
%\todols{We need to introduce the idea of evaluating these metrics (DA, etc.) as a ranking problem - this is not said anywhere and then then mentions of ranking below come out of the blue. Say: "We propose to assess the usefulness of metrics according to their ability to rank translations based on the time that would be required to post-edit them. This has a very practical application in the translation industry.... - Carol to continue.}
%%%%%%%%%%%%%%%%%%%%

%%%%%%%%%%%%%%%%%%%%%%%%%%%%%%%%%%%%%%%%%%%%%%%%%%%%%%%%%%%%%%%%%%%%%%%%
\begin{table*}[ht]
    \centering
   \scalebox{0.8}{
%    \scalebox{0.9}{
    \begin{tabular}{l|c|c|c|c|c|c|c|c|c|c|c|c}
        & \multicolumn{2}{c|}{ANN0} & \multicolumn{2}{c|}{ANN1} & \multicolumn{2}{c|}{ANN2} & \multicolumn{2}{c|}{ANN3} & \multicolumn{2}{c|}{ANN4} & \multicolumn{2}{c}{ALL}\\
        & Mean & st.\ dev. & Mean & st.\ dev. & Mean & st.\ dev. & Mean & st.\ dev & Mean & st.\ dev & Mean & st.\ dev \\
        \hline
        HTER &0.32 & 0.17 &  0.27 & 0.25 &  0.25 & 0.16 &  0.30 & 0.18 &  0.30 & 0.18 & 0.29 & 0.19\\
        HBLEU & 0.49 & 0.21 &  0.60 & 0.26 &  0.57 & 0.21 &  0.52 & 0.22 &  0.53 & 0.21 & 0.54 & 0.23 \\
        HMETEOR & 0.65 & 0.16 &  0.72 & 0.25 &  0.72 & 0.16 &  0.67 & 0.17 &  0.68 & 0.16 & 0.69 & 0.19  \\
        \hline
        Keys/char & 0.43 & 0.33 &  0.44 & 0.40 &  0.46 & 0.41 &  0.55 & 0.42 &  0.42 & 0.35 & 0.46 & 0.39 \\
        \hline
        \hline
        \petpw{} (sec/word) & 3.88 & 2.91 & 2.42 & 2.78 & 3.66 & 2.71 & 3.58 & 2.31 & 4.23 & 4.22 & 4.23 & 4.22 \\
    \end{tabular}}
    \caption{Statistics (mean and standard deviation) of task-specific (PE-based) metrics in our dataset}
    \label{tab:stats}
\end{table*}
%%%%%%%%%%%%%%%%%%%%%%%%%%%%%%%%%%%%%%%%%%%%%%%%%%%%%%%%%%%%%%%%%%%%%%%%

Our main contributions are:
\begin{itemize}\itemsep 0ex
    %\item a study of DA and average PE time per word (\petpw{}), %\footnote{PE time refers to the normalised version of PE time per word.\todols{confusing: say instead: PE time is normalised as the average PE per word?}} 
    %showing that they behave very differently;
%    \todomlf{Do we really show which aspects are evaluated by each metric?}
%    \todocs{We show they are different and we assume PE time is evaluating PE effort, so DA is not a good proxy for PE effort.}
%    \todomlf{But *what* other aspects does DA measure (if any)?}
    \item a comprehensive review of task-specific (PE-based) metrics (e.g.\ HTER), reference-based metrics (e.g.\ TER) and DA, where the goal is to rank \mted{} segments according to PE time; %\todols{I'm not sure we need contribution 1 - 1 and 2 are the same, aren't they? } % \petpw{} is our gold standard;
    %\item a critique of previous work that has blindly relied on DA for measuring PE effort, showing that this is not a valid assumption;
    \item the release of a dataset with source, \mted{}, reference, and \peed{} texts; detailed information about five independent post-editing jobs for each \mted{} text, and DA annotations;\footnote{\url{https://github.com/carolscarton/iwslt2019}}
    \item a new ranking score for \mted{} segments called \emph{split-averaged, time-ratio assessment} (SATRA).
    %\todocs{I added a footnote in section 2. Tried to add here, but for some reason the overleaf crashed.}%\footnote{We will add a link to the corpus after the paper is accepted.}
\end{itemize}

% Adapt to new version of paper
In Section \ref{sec:data} we present the dataset created and used for this paper. 
% What happens if we remove this section?
%Section \ref{sec:pevsda} shows an analysis of DA versus PE time.  
%
Section \ref{sec:analysis} presents our ranking analysis using all evaluation metrics available in our dataset. 
%Section \ref{sec:pearson} discusses the use of Pearson's $r$ correlation scores in order to assess correlation between evaluation metrics and how this approach can result in misleading conclusions. 
In section \ref{sec:related} we discuss related work. The paper ends with concluding remarks (Section~\ref{sec:remarks}). %\todols{you still don't describe the content of section 5 :-) - not sure how to introduce it} %\todols{strange place for related work - move as section 2 or last before conclusions}.

\section{Dataset and annotators}\label{sec:data}

We extend the dataset made available by the WMT 2016 shared task on document-level quality estimation \cite{Bojar2016}. This dataset contains 1,047 segments totalling 26,875 words, \mted{} by 41 different systems ---with an average of 26 segments per system---
extracted from the test sets of English--Spanish WMT translation shared tasks between 2008 and 2012.  Existing \mted{} segments were crowd-annotated (via Amazon Mechanical Turk) using DA scores made available by Graham et al.~\cite{Graham-etal:2017}.\footnote{\url{https://github.com/ygraham/eacl2017}}
%\todomeg{Probably it is here where we should explain that we used EXACTLY the same DA data from the paper by Graham et al., but I am not sure how to explain this without breaking anonymity}

%\todols{Here I had meant the profile of our annotators - it suffices to say we rely on professional translators with experience in PE, I don't think we need to say anything about the DA data as we say it was crowdsourced}
%\footnote{The profile of annotators is not described. Control measures were however embedded in the job.}
%\todols{Also, give profile of annotators (professionals? experience post-editors?}
Although the aim of the work presented in~\cite{Graham-etal:2017} was to generate DA scores at the document level, they first assessed each segment independently. Each segment has then a DA score and these are the values used in our experiments. The DA value of each segment is obtained by %getting the mean score over 
averaging
the assessment of various annotators (previous work recommend at least $15$ annotations per segment \cite{graham-etal:2015}, and the study in~\cite{Graham-etal:2017} follows the same protocol). 
%\todols{we should say that DA is computed as the z-normalised average of X crowd annotations}\todomlf{Added later}

For the post-editing task, we hired five professional translators with experience in PE (hereafter referred to as ANN0, ANN1, ANN2, ANN3 and ANN4), who generated \peed{} versions for each segment of this dataset. The annotators used the PET tool~\cite{Aziz-etal:2012}, which records the 
%details of
edit operations performed during the PE task, including the time elapsed to post-edit a 
segment. %segment (as explained in Section \ref{sec:pevsda}, 
We use PE time normalised by the length of the target segment in words, \petpw{}). 

We then calculated the following task-specific (PE-based) metrics: 
\begin{itemize}\itemsep 0ex
    \item HTER, HBLEU and HMETEOR, respectively the TER, BLEU \cite{Papineni2002}, and METEOR \cite{Banerjee2005} scores of the \mted{} segment using the \peed{} version as reference,\footnote{These metrics were calculated using the Asiya toolkit \cite{PBML_Asiya:2010}.} and
    \item Keys/char: ratio between the number of keys pressed by an annotator and the number of characters in the \mted{} segment. %\todols{is it the MT or PE segment?}\todomlf{\mted{} segment, confirmed}.
\end{itemize}

Since we have access to the references from the WMT datasets, we also calculated  standard reference-based BLEU, METEOR and TER scores.

Table \ref{tab:stats} shows some statistics of the task-specific (PE-based) metrics extracted for this dataset. We show statistics per annotator and also 
% considering 
the averaged values 
%of
for
all annotators (ALL). Statistics for DA, and reference-based metrics are shown in Table \ref{tab:stats-da-ref}. All averages in both tables are weighted by the number of \mted{} words, as post-editing time ---the measurement we want to track--- is expected to grow linearly with sentence length.
%\todomeg{Should we mention the range of values that DA, TER, BLEU and METEOR can take in table 3?}\todomlf{Not sure. Some are not bound}
As may be seen, the values of quality indicators show a rather wide range.

\begin{table}[ht]
    \centering
    \scalebox{0.9}{
    %\scalebox{0.8}{
    \begin{tabular}{l|c|c}
       
        & Mean & st.\ dev. \\
        \hline
        DA & -0.02 & 0.61\\
        \hline
        TER & 0.57 & 0.21 \\
        BLEU & 0.24 & 0.16\\
        METEOR & 0.42 & 0.16 \\
    \end{tabular}}
    \caption{Statistics (mean and standard deviation) of DA and reference-based metrics in our dataset}
    \label{tab:stats-da-ref}
\end{table}

\begin{table*}[]
    \centering
%    \scalebox{0.8}{
   \scalebox{0.9}{
    \begin{tabular}{l|c|c|c|c|c|c|c|c|c|c|c|c}
            & \multicolumn{2}{c|}{ANN0} & \multicolumn{2}{c|}{ANN1} & \multicolumn{2}{c|}{ANN2} & \multicolumn{2}{c|}{ANN3} & \multicolumn{2}{c|}{ANN4} & \multicolumn{2}{c}{ALL} \\
            & $\rho$ & S & $\rho$ & S & $\rho$ & S & $\rho$ & S & $\rho$ & S & $\rho$ & S\\
            \hline

            TER & .24* & .78 & .32* & .67 & .26 & .73 & .23~ & .81 & .20~ & .83 & .30 & .77 \\
            BLEU & .25* & .74 & .33* & .64 & .29 & .70 & .30* & .75 & .23* & .77 & .33 & .72 \\
            METEOR & .25* & .74 & .34~ & .63 & .31 & .67 & .30* & .76 & .23* & 75 & .35 & .71
           
            \\ \hline
                        DA & .38 & .68 & .48 & .59 & .44 & .66 & .45 & .70 & .43 & .62 & .52 & .64 \\
            
            \hline
            HTER & .58~ & .53 & .62~ & .47 & .71 & .47 & .67 & .54 & .61~ & .49 & .69 & .53 \\
            HBLEU & .54* & .54 & .60* & .49 & .67 & .48 & \textbf{.68} & .54 & .58* & .50 & .68 & .53 \\
            HMETEOR & .53* & .55 & .61* & .48 & .69 & .47 & .65 & .54 & .59* & .50 &  .68 & .54\\
            \hline
            Keys/char & \textbf{.63} & \textbf{.48} & \textbf{.75} & \textbf{.37} & \textbf{.74} & \textbf{.45} & \textbf{.68} & \textbf{.52} & \textbf{.63} & \textbf{.43} & \textbf{.76} & \textbf{.49}\\
            \hline
            \hline
            \petpw{} & 1.0 & .31 & 1.0 & .25 & 1.0 & .32 & 1.0 & .38 & 1.0 & .26 & 1.0 &  .39  \\
    \end{tabular}}
    \caption{Spearman's $\rho$ ($\uparrow$) and SATRA (S, $\downarrow$) scores for all metrics using \petpw{} as gold standard. The best results are shown in bold and * means no statistically significant difference between the metrics according to Williams test with $p < 0.01$.}
    \label{tab:ranking-analysis}
\end{table*}

\begin{table*}[]
    \centering
    %\scalebox{.8}{
    \scalebox{0.9}{
    \begin{tabular}{l|c|c|c|c|c|c|c|c|c|c}
            & \multicolumn{2}{c|}{ANN0} & \multicolumn{2}{c|}{ANN1} & \multicolumn{2}{c|}{ANN2} & \multicolumn{2}{c|}{ANN3} & \multicolumn{2}{c}{ANN4} \\
            & $\rho$ & S & $\rho$ & S & $\rho$ & S & $\rho$ & S & $\rho$ & S \\
            \hline
            DA & .52 & .63 & .51 & .65 & .51 & .64 & \textbf{.61} & .64 & .52 & .65 \\
            \hline
            HTER & \textbf{.59} & .59 &.45 & .71 &\textbf{.60} & \textbf{.58} & .57 & .59 & \textbf{.62} & \textbf{.57} \\
            HBLEU & .57 & .59 & .45 & .73 & .57 & .59 & .56 & .60 & .60 & .58 \\
            HMETEOR & .57 & .59 & .42 & .72 & .58 & .59 & .55 & .60 & .60 & .58 \\
            \hline
            Keys/char & \textbf{.59} & \textbf{.58} & \textbf{.54} & \textbf{.62} & .57 & .60 & .59 & \textbf{.58} & .60 & .58 \\
            \hline
            \hline
            \petpw{} & .58 & .53 & .62 & .57 &.61 & .57 & .62 & .55 & .63 & .55\\
    \end{tabular}}
    \caption{Spearman's $\rho$ ($\uparrow$) and SATRA ($\downarrow$) scores for all metrics using \petpw{} as gold standard for the leave-one-out experiment}
    \label{tab:ranking-leave-one-out}
\end{table*}

%%%%%%%%%%%%%%%%%%%%%%%%%%%%%%%%%%%%%%%%%%%%%%%%%%%%%%%%%%%%%%%%%%%%%%%%%%

%\section{Analysis of rankings} 
\section{Comparing the ranking ability of metrics}
\label{sec:analysis}

In order to analyse %which evaluation metrics can be used as 
the performance of evaluation metrics as 
a proxy for \petpw{}, we propose experiments that look at how these metrics rank \mted{} segments. We argue that looking at the rankings gives a  reliable perspective of the usefulness of the metrics for the PE task, since it gives us the relative differences, in terms of effort, among the segments to be \peed{}.%\todomlf{Check last sentence please!}

%%%%%%%%%%%%%%%%%%%%%%
%\todols{important: this explanation is not true for ranking - this are general motivations for predictive quality (i.e. QE), not for ranking according to PE time (PE would be needed anyway to measure that). Revise: do you want to say that we can use PE time to build prediction models that can then rank translations? even so, the definition above is true for binary prediction more so than for ranking. Maybe talk about ranking as a 'tool' to analyse the metrics, not as a final application/task}
%%%%%%%%%%%%%%%%%%%%%%%%%%
%Ranking machine translated segments is particularly desirable when the focus is to estimate productivity. From the perspective of the post-editor, ranking translations according to estimated PE time per word would allow them to decide which segments are going to take longer to post-edit or even if post-editing is worthwhile. Companies, interested in reducing costs, may also make use of segments ranked by quality.

\subsection{Ranking correlation}

In this experiment, we try to identify which metric produces rankings that are closest to \petpw{} rankings. Firstly, we calculate Spearman's $\rho$ rank correlation coefficient between \petpw{} and all metrics. In addition to $\rho$, we also compute a new ranking score called  \emph{split-averaged, time-ratio assessment} (SATRA) for a ranking \(R\) as follows:
\[
\mathrm{SATRA}(R)=\frac{1}{N-1}\sum_{j=1}^{N-1}
\frac{\tau_1^j(R)}{\tau_{j+1}^N(R)}
\]
with 
%\[\tau_m^n(R)=\frac{\displaystyle\sum_{i=m}^n T(R_j)}{\displaystyle\sum_{i=m}^n L(R_j)},\]
\[\tau_m^n(R)=\sum_{j=m}^n T(R_j)\left(\displaystyle\sum_{i=m}^n L(R_j)\right)^{-1},\]
the average measured \petpw{} for segments \(R_m\) to \(R_n\) (those ranked $m$-th to $n$-th), where \(T(R_j)\) is the total PE time and \(L(R_j)\) the total length in (\mted{}) words for segment \(R_j\) (ranked $j$-th). The value of \(\mathrm{SATRA}(R)\) should be close to 1 for a random ranking (the  average \petpw{} above any split of the ranking and that below the split should roughly be the same), smaller than 1 for a good ranking (one that would rank easier-to-post-edit segments better than hard-to-post-edit ones), and the minimum possible for a ranking based on the measured \petpw{}.
These two scores are used to measure how close two ranked distributions are.\footnote{SATRA is similar to DeltaAVG \cite{callisonburch-EtAl:2012:WMT} but has a simpler interpretation in terms of the average PE time per word above and below any split of the rank.} %\todols{say this is similar to DeltaAVG (WMT12), but better because?}

Table \ref{tab:ranking-analysis} shows Spearman's $\rho$ correlation coefficients and SATRA scores between all metrics and the individual \petpw{} of all annotators and the averaged values of all annotators (ALL). The last line of the table provides the scores obtained by an oracle using the actual \petpw{} as the ranking metric; this helps to interpret SATRA scores as, unlike Spearman's $\rho$, they do not have a fixed lower bound. %Although 
DA shows moderate Spearman's $\rho$ values across all annotators and for ALL, 
%they 
which are considerably smaller than those achieved by HTER, HBLEU, HMETEOR and Keys/char. SATRA shows similar results: DA presents larger (worse) values than the task-specific PE-based metrics. Following previous work~\cite{graham:2015}, we calculate the statistical significance difference between all metrics using Williams' test over the Spearman's $\rho$ scores ($p<0.01$).\footnote{Williams test is calculated using  \texttt{mt-qe-eval}: \url{https://github.com/ygraham/mt-qe-eval}.} The large majority of the results are statistically different.

The best overall metric is Keys/char, %achieving 
which achieves
the highest Spearman's $\rho$ scores and the lowest SATRA scores for all annotators individually and for ALL. The only annotator where the Keys/char metric is not so salient is ANN3. Our hypothesis is that this annotator may have interacted more with the mouse,\footnote{PET records keyboard actions, but not mouse actions.} instead of with the keyboard. HTER, HBLEU and HMETEOR do not show significant differences among them. This is in line with the results reported by previous work~\cite{Graham-etal:2016} that found no difference between these metrics when correlating them to DA. 
%Therefore, they seem to be measuring the same aspect of quality. 
Finally, independent-reference-based metrics show the worst ranking scores with respect to \petpw{}.

\begin{figure*}[h]
  \includegraphics[width=\textwidth]{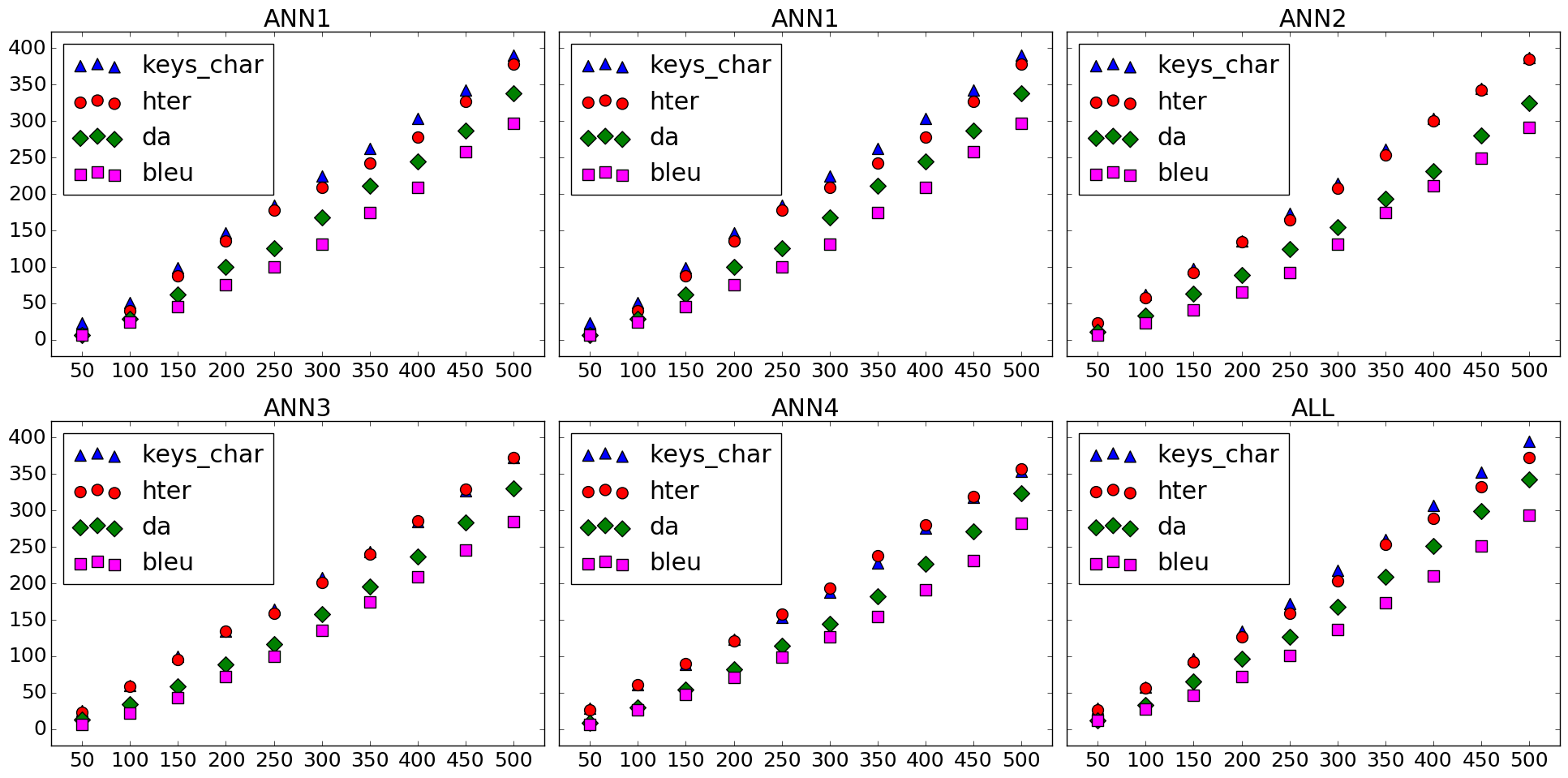}
  \caption{Number of segments shared between the 500 best sentences acording to \petpw{} and the other metrics}
  \label{fig:best}
\end{figure*}
%%%%%%%%%%%%%%%%%%%%%%%%%%%%%%%%%%%%%%%%%%%%%%%%%%%%%%%%%%%%%%

In a real-world scenario, the \petpw{} of one annotator could be estimated based on the \petpw{} of other annotator(s). In order to simulate this case and evaluate whether the results from Table~\ref{tab:ranking-analysis} would still stand, we performed leave-one-out experiments. In this case, SATRA and Spearman's $\rho$ scores are calculated between each one of the studied metrics for one annotator and the averaged \petpw{} of all other annotators. For example, for ANN0, the \petpw{} is the average \petpw{} of ANN1 to ANN4, and its correlation with DA and the HTER, HBLEU, HMETEOR, Keys/char and \petpw{} for ANN0 post-edits. Table~\ref{tab:ranking-leave-one-out} shows the results of this experiment. As expected, the difference between Spearman's $\rho$ and SATRA scores for HTER, HBLEU and HMETEOR and for Keys/char is lower than in Table~\ref{tab:ranking-analysis}, since we are not dealing with the individual \petpw{} of each annotator.
SATRA scores for PE-based metrics are better (lower) than DA (except for ANN1), and similarly for Keys/char (except for ANN1).  In addition, Keys/char is not the best metric overall anymore, although it still shows the best SATRA in three out of five cases.  In general, PE-based approaches still outperform DA in most cases. For reference, the last row of Table~\ref{tab:ranking-leave-one-out} shows Spearman's $\rho$ and SATRA for the \petpw{} of each annotator versus the leave-one-out \petpw{}.

%\begin{table*}[]
%    \centering
%    \scalebox{0.9}{
%    \begin{tabular}{l|c|c|c|c|c|c|c|c|c|c}
%            & \multicolumn{2}{c|}{ANN0} & \multicolumn{2}{c|}{ANN1} & \multicolumn{2}{c|}{ANN2} & \multicolumn{2}{c|}{ANN3} & \multicolumn{2}{c}{ANN4} \\
%            & D & S & D & S & D & S & D & S & D & S \\
%            \hline
%            DA & 0.999 & 0.52 & \textbf{1.040} & 0.51 & 0.984 & 0.51 & 1.005 & \textbf{0.61} & 0.925 & 0.52 \\
%            \hline
%            HTER & 1.053 & \textbf{0.59} & 0.881 & 0.45 & \textbf{1.121} & \textbf{0.60} & 1.037 & 0.57 & 1.052 & \textbf{0.62}\\
%            HBLEU & 1.047 & 0.57 & 0.880 & 0.45 & 1.113 & 0.57 & 1.086 & 0.56 & 1.010 & 0.60\\
%            HMETEOR & \textbf{1.082} & 0.57 & 0.863 & 0.42 & 1.099 & 0.58 & 1.060 & 0.55 & \textbf{1.067} & 0.60 \\
%            \hline
%            Keys/char & 1.072 & \textbf{0.59} & 1.033 & \textbf{0.54} & 1.074 & 0.57 & \textbf{1.115} & 0.59 & 1.033 & 0.60 \\
%            \hline
%            \hline
%            PE time & 1.129 & 0.58 & 1.264 & 0.62 & 1.201 & 0.61 & 1.253 & 0.62 & 1.198 & 0.63\\
%    \end{tabular}}
%    \caption{DeltaAVG ($\uparrow$) and Spearman's $\rho$ ($\uparrow$) scores for all metrics using PE time as gold standard for the leave-one-out experiment}
%    \label{tab:ranking-leave-one-out}
%\end{table*}

It is worth mentioning that, with only five annotators, it is difficult to devise a model that would be a good estimator of quality for new annotators. In fact, after doing an analysis using the distribution-agnostic Kolmogorov--Smirnov
 test\footnote{\protect{\url{https://en.wikipedia.org/wiki/Kolmogorov-Smirnov_test}}}
over the \petpw{} distributions (considering $p<0.05$), 
we identified three clusters of annotators where their \petpwh{} measurements come from the same distribution. 
Basically, ANN0, ANN2 and ANN4 could be clustered together, whilst ANN1 and ANN3 would have their own clusters. This may be impacting our results, but a deeper analysis of the effect of such clusters is left for future work.

\subsection{Analysis of tails}
%\todomlf{This is also quantitative as the first section!}

The experiments in this section aim to %
%emphasise that DA and \petpw{} do not measure the same aspects of quality and, therefore, DA should not be used as a proxy for \petpw{} or for estimating the effort or cost of post-editing.
obtain a closer view of how the metrics studied perform for the best and the worst segments, 
%\todols{this red bit is repetition of what was said before, but possibly said better here. In any case, it's too late for it, it should appear early, not here}
%Simulating this scenario, we propose 
by performing
an analysis of the tails of the \petpw{} distribution. In other words, we want to analyse how the task-specific PE-based metrics, reference-based metrics, and DA perform on the best and worst segments according to \petpw{}. Our experiment consists in counting the number of common segments between different cuts of the \petpwh{} ranking and the rankings obtained with each different metric. 

%%%%%%%%%%%%%%%%%%%%%%%%%%%%%%%%%%%%%%%%%%%%%%%%%%%%%%%%%%%%%%%%%%%%
%\begin{table*}[ht]
%    \centering
%%    \scalebox{0.8}{
%    \scalebox{0.9}{
%    \begin{tabular}{l|c|c|c|c|c|c}
%            & ANN0 & ANN1 & ANN2 & ANN3 & ANN4 & ALL \\
%            \hline
%            DA & -0.901 & -0.769 & -0.785 & -0.858 & -0.793 & -0.894 \\
%            \hline
%            BLEU & -0.647 & -0.488 & -0.554 & -0.614 & -0.520 & -0.651\\
%            TER & 0.881 & 0.759 & 0.787 & 0.850 & 0.784 & 0.879\\
%            METEOR & -0.791 & -0.639 & -0.678 & -0.749 & -0.668 & -0.789\\
%            \hline
%            HTER & 0.908 & 0.791 & 0.819 & 0.872 & 0.821 & 0.911 \\
%            HBLEU & -0.776 & -0.700 & -0.698 & -0.740 & -0.667 & 0.819\\
%            HMETEOR & -0.869 & -0.777 & -0.783 & -0.833 & -0.770 & 0.777\\
%            \hline
%            Keys/char & \textbf{0.960} & \textbf{0.953} & \textbf{0.945} & \textbf{0.938} & \textbf{0.907} & \textbf{0.950}\\
        
%    \end{tabular}}
%    \caption{Pearson's $r$ correlation scores between \petpw{} and all metrics}
%    \label{tab:pearson}
%\end{table*}
%%%%%%%%%%%%%%%%%%%%%%%%%%%%%%%%%%%%%%%%%%%%%%%%%%%%%%%%%%%%%%%%%%%%%

\textbf{Best segments:} firstly, we look at the first 500 sentences in the \petpwh{} ranking, that is, the 500 easiest-to-post-edit sentences, and compare to the first 500 sentences in the rankings according to other metrics. We split the rankings in sets of 50 to show the performance of the metrics and the differences among them. Figure \ref{fig:best} shows the results of this experiment for all annotators individually and for the ALL case. For clarity we only show four metrics: Key/char, HTER, DA and BLEU. One can clearly identify three groups of metrics:
\begin{itemize}\itemsep 0ex
    \item BLEU, TER and METEOR rankings behave similarly and show the lowest number of segments in common to the \petpwh{} ranking;
    \item HTER, HBLEU, HMETEOR, and Keys/char rankings are the best, sharing the largest number of segments with the \petpwh{} ranking;
    \item DA ranking is better than the reference-based metrics, but worse than the task-specific PE-based metrics.
\end{itemize}
These findings are in agreement with those obtained when ranking all segments.

\begin{figure}[h]
\centering

 \includegraphics[width=0.49\textwidth]{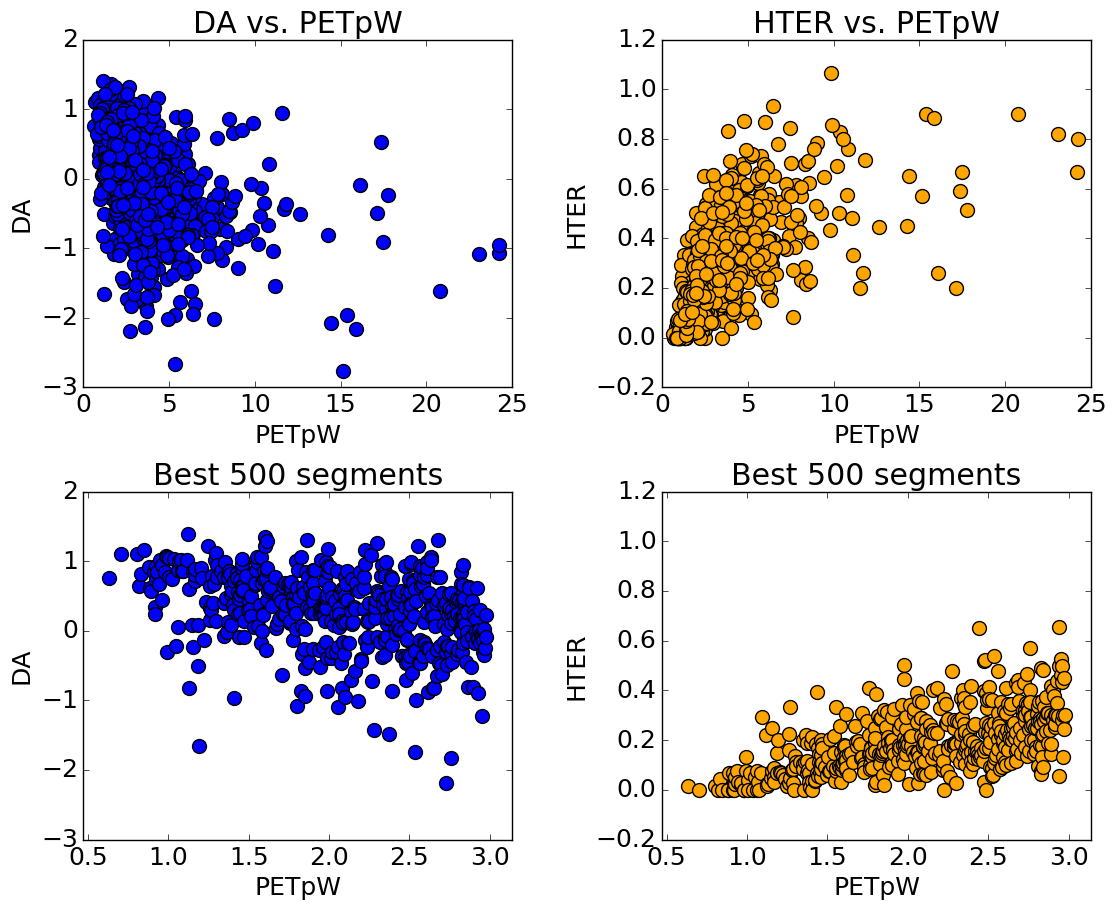}
  \caption{Scatter plots for DA vs.\ \petpw{} and HTER vs.\ \petpw{}}
  \label{fig:scatter}
\end{figure}

Figure \ref{fig:scatter} shows scatter plots for DA vs.\ \petpw{} and HTER vs.\ \petpw{} averaged over all five annotators. The top two graphs show the scatter plots for the entire dataset. In this case, both metrics look similar in comparison to \petpw{}, although DA seems to show more outliers. The bottom two graphs show the scatter plots for the best 500 segments according to \petpw{}. In this case, HTER shows a clear tendency, where the majority of the values have a low HTER score and a low \petpw{}. DA, on the other hand, shows a much sparser graph.

\begin{figure}[ht]
\centering
  \includegraphics[width=0.48\textwidth]{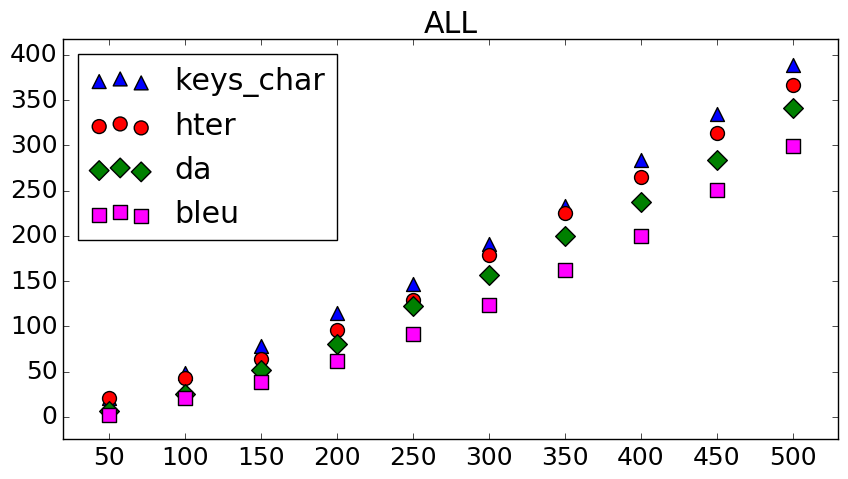}
  \caption{Number of segments shared between the $500$ worst sentences according to \petpw{} and the other metrics for all translators.}
  \label{fig:worst}
\end{figure}

\textbf{Worst segments:} a similar trend is shown when we analyse the 500 worst segments (due to space constraints, Figure \ref{fig:worst} only shows results for ALL), although the gap between DA and task-specific PE-based metrics is smaller. One hypothesis is that, for the worst segments, where the quality is very low, differences in adequacy track differences in PE time better.
%more salient aspect of quality for the post-editing task. 
%performance of DA is consistently below that of task-based ce of DA is consistently below that of tas\left\(based metrics. 

%\section{Discussion}

\section{Related work}\label{sec:related}

%Machine-translated texts may be used for different applications as discussed in the previous section. 
%As a consequence, several ways of assessing MT quality have been proposed. %, focusing on different purposes. 
%Reference-based metrics (e.g. BLEU) sometimes fail to assess MT usefulness for a specific task, such as estimating PE effort. \todomlf{Look for a reference for this?} Our hypothesis is that DA scores are also not ideal for this purpose since this judgment-based score only assesses general adequacy. 
In what follows we present previous work on human task-based evaluation that targeted PE effort and on the use of DA for the same purpose.%\todomlf{Text that was really not needed has been removed. Please check.}

\textbf{PE time} is a straightforward indicator of MT quality: segments that take longer to be \peed{} are considered worse than segments that can be quickly corrected. Koponen et al.\ \cite{Koponen2012a} argue that PE time is the most effective way of measuring cognitive aspects of the PE task and
%relating
relate
them to the quality of the translations. Plitt and Masselot \cite{Plitt2010} use \petpw{} (actually, its converse: words per hour) to measure the gain in productivity when post-editing \mted{} text ---in a real translation workflow--- over the productivity when performing translation from scratch.

% can be shortened
\textbf{Perceived PE effort:} humans are asked to give a score for the \mted{} sentences according to a
% Likert is a researcher's name and does not need italics
%\textit{Likert}
Likert
\cite{likert1932technique} scale representing perceived PE effort \cite{Specia2011}. This type of score can be given with or without actual post-editing and it represents a judgement on how difficult it would be (or it was) to fix the given \mted{} sentence. Perceived PE effort scores were used in the WMT 2012 \cite{callisonburch-EtAl:2012:WMT} and WMT 2014 \cite{Bojar2014} QE shared task editions.
 
\textbf{Eye-tracking:} previous work have also relied on eye-tracking to evaluate PE effort. O'Brien \cite{OBrien2011} measures fixation time and correlates it with GTM (a similarity metric between the machine translation and the reference sentence based on precision, recall and $F$-measure \cite{Turian2003}). Low GTM scores show correlation with high fixation time. PE pauses (extracted from keystroke logs) can also be viewed as an indirect measure of cognitive effort \cite{Lacruz2014}. Long pauses are associated with segments that demand more cognitive PE effort.

\textbf{Edit distance and $n$-gram-based scores:} PE effort can also be evaluated indirectly, by using a metric that takes into account edit operations. HTER \cite{Snover-etal:2006} is an example of such a metric, which computes the minimum number of edits to transform the machine translation into the \peed{} version. Task-specific, PE-based (\emph{human} or \emph{H-}) variants of commonly-used reference-based similarity measures have also been studied, such as HBLEU and HMETEOR. However, HTER is the most widely used as an indirect measurement of PE effort \cite{Bojar2016,Bojar2014,Bojar2013,Bojar2015,Bojar2017,Specia2018}

%\todomlf{Try a less-confrontational approach}
\textbf{DA:} Graham et al. \cite{Graham-etal:2014,Graham-etal:2016-da} propose the use of DA for MT evaluation. According to the authors, the biggest advantage of their approach in comparison to early practices of adequacy judgements is that they can reliably crowd-source the annotations. 
%, being able to acquire many annotations for each segment and alleviate the problem with annotator's biases. 
Graham et al.\ \cite{Graham-etal:2016} also express a strong criticism of HTER on the grounds that it does not show high Pearson $r$ correlation scores with DA. In another work~\cite{Graham-etal:2017}, the same authors also criticise a variant of HTER for document-level QE, suggesting that DA is a more adequate metric to compare different QE systems. 
Recently, Bentivogli et al.\ \cite{Bentivogli-etal:2018} evaluate HTER and mTER (multi-reference TER) against DA scores and conclude that mTER is a better proxy for PE effort because it shows higher correlation scores with DA than HTER. However, our analysis on a real-world measurement of productivity (\petpw{}) show that PE-based metrics (including HTER) are the most adequate metrics to approximate \petpw{}, outperforming DA. Therefore, we argue that if mTER and HTER were compared using their correlations to \petpw{}, the results could be different (this analysis is left for future work). 

%\todols{this is the only really related work - we need to say how we are different from it...}
%\todomlf{Shouldn't we be a bit more critical of the DA %\todols{yes, we need to close this section with that}

\section{Concluding remarks} \label{sec:remarks}
%\todomlf{First attempt --- basically echoes the abstract}

The advancement and adoption of MT depends more than ever on the availability of reliable metrics to evaluate its quality. Averaged subjective \emph{direct assessment} (DA) of MT quality, which is performed independently of purpose and may easily be crowd-sourced, has become very popular.
%, and has even been used as gold standard to assess MT evaluation metrics in evaluation campaigns such as WMT17 \cite{bojar-graham-kamran:2017:WMT} and WMT18 \cite{ma-bojar-graham:2018:WMT}. 
However, in an important application of MT, namely dissemination via post-editing, it is only natural to use actual \emph{measurements} that are obtained after performing post-editing. It is also natural for quality estimation models to target such metrics. 

%It is worth mentioning that DA is a better estimator of \petpw{} than reference-based metrics and that PE-based metrics are more expensive, cumbersome and not always easily available. However, QE techniques enable training models capable of predicting task-based metrics using a small number of data points.

The results of our experiments on a dataset that includes PE indicators collected for five translators show that DA \emph{judgements} provide a reasonable approximation 
%are not the most adequate estimate 
of relevant, measurable aspects of MT usefulness in a dissemination task, such as PE time;
%
%they also show that, 
however,
as expected, task-specific metrics comparing \mted{} and \peed{} text -- such as HTER or the number of keystrokes per raw MT character -- are 
%much 
better 
%predictors
trackers
of \petpw{}. 
%added
DA does however perform better than metrics such as BLEU, TER or METEOR with respect to an independent reference translation.

These results lead us to recommend that MT practitioners should use task-specific metrics wherever this is possible, and non-expert subjective %DA
judgements such as DA 
only when specific, measurable metrics are not available or feasible for a task.

\textbf{Acknowledgements:} Work supported by the Spanish government
through project EFFORTUNE (TIN2015-69632-R) and  grant PRX16/00043
for MLF, and by the European Commission through project GoURMET (No.\ 825299). CS and MLF are both first authors with equal contribution.
\bibliography{wmt19}

\begin{thebibliography}{10}
\providecommand{\url}[1]{#1}
\csname url@rmstyle\endcsname
\providecommand{\newblock}{\relax}
\providecommand{\bibinfo}[2]{#2}
\providecommand\BIBentrySTDinterwordspacing{\spaceskip=0pt\relax}
\providecommand\BIBentryALTinterwordstretchfactor{4}
\providecommand\BIBentryALTinterwordspacing{\spaceskip=\fontdimen2\font plus
\BIBentryALTinterwordstretchfactor\fontdimen3\font minus
  \fontdimen4\font\relax}
\providecommand\BIBforeignlanguage[2]{{%
\expandafter\ifx\csname l@#1\endcsname\relax
\typeout{** WARNING: IEEEtran.bst: No hyphenation pattern has been}%
\typeout{** loaded for the language `#1'. Using the pattern for}%
\typeout{** the default language instead.}%
\else
\language=\csname l@#1\endcsname
\fi
#2}}

\bibitem{Nirenburg:1993}
S.~Nirenburg, \emph{Progress in Machine Translation}.\hskip 1em plus 0.5em
  minus 0.4em\relax Amsterdam, Netherlands: IOS B. V., 1993.

\bibitem{sanchez2016machine}
\BIBentryALTinterwordspacing
M.~Sanchez-Torron and P.~Koehn, ``Machine translation quality and post-editor
  productivity,'' in \emph{Proceedings of the Conference of the Association for
  Machine Translation in the Americas Vol. 1: MT Researchers' Track}, Austin,
  TX, 2016, pp. 16--26. [Online]. Available:
  \url{https://amtaweb.org/wp-content/uploads/2016/10/AMTA2016_Research_Proceedings_v7.pdf}
\BIBentrySTDinterwordspacing

\bibitem{hassan2018}
\BIBentryALTinterwordspacing
H.~Hassan, A.~Aue, C.~Chen, V.~Chowdhary, J.~Clark, C.~Federmann, X.~Huang,
  M.~Junczys{-}Dowmunt, W.~Lewis, M.~Li, S.~Liu, T.~Liu, R.~Luo, A.~Menezes,
  T.~Qin, F.~Seide, X.~Tan, F.~Tian, L.~Wu, S.~Wu, Y.~Xia, D.~Zhang, Z.~Zhang,
  and M.~Zhou, ``Achieving human parity on automatic chinese to english news
  translation,'' \emph{Computing Research Repository}, vol. arXiv:1803.05567,
  2018. [Online]. Available: \url{http://arxiv.org/abs/1803.05567}
\BIBentrySTDinterwordspacing

\bibitem{laubli2018}
\BIBentryALTinterwordspacing
S.~L{\"a}ubli, R.~Sennrich, and M.~Volk, ``Has machine translation achieved
  human parity? a case for document-level evaluation,'' in \emph{Proceedings of
  the 2018 Conference on Empirical Methods in Natural Language
  Processing}.\hskip 1em plus 0.5em minus 0.4em\relax Association for
  Computational Linguistics, 2018, pp. 4791--4796. [Online]. Available:
  \url{http://aclweb.org/anthology/D18-1512}
\BIBentrySTDinterwordspacing

\bibitem{toral-etal:2018}
\BIBentryALTinterwordspacing
A.~Toral, S.~Castilho, K.~Hu, and A.~Way, ``{Attaining the Unattainable?
  Reassessing Claims of Human Parity in Neural Machine Translation},'' in
  \emph{Proceedings of the Third Conference on Machine Translation (WMT),
  Volume 1: Research Papers}.\hskip 1em plus 0.5em minus 0.4em\relax Brussels,
  Belgium: Association for Computational Linguistics, 2018, pp. 113--123.
  [Online]. Available: \url{http://www.statmt.org/wmt18/pdf/WMT012.pdf}
\BIBentrySTDinterwordspacing

\bibitem{Krings2001}
H.~P. Krings, \emph{Repairing texts: Empirical investigations of machine
  translation post-editing process}.\hskip 1em plus 0.5em minus 0.4em\relax
  Kent, OH: The Kent State University Press, 2001.

\bibitem{snover2008terp}
M.~Snover, N.~Madnani, B.~Dorr, and R.~Schwartz, ``{TERp} system description,''
  in \emph{Proceedings of the MetricsMATR workshop}, vol.~34, no.~67, 2008, p.
  108.

\bibitem{Snover-etal:2006}
M.~Snover, B.~Dorr, R.~Schwartz, L.~Micciulla, and J.~Makhoul, ``{A Study of
  Translation Edit Rate with Targeted Human Annotation},'' in \emph{Proceedings
  of the Seventh biennial conference of the Association for Machine Translation
  in the Americas}, Cambridge, MA, 2006, pp. 223--231.

\bibitem{Graham-etal:2016}
\BIBentryALTinterwordspacing
Y.~Graham, T.~Baldwin, M.~Dowling, M.~Eskevich, T.~Lynn, and L.~Tounsi, ``Is
  all that glitters in machine translation quality estimation really gold?'' in
  \emph{Proceedings of the 26th International Conference on Computational
  Linguistics: Technical Papers}, Osaka, Japan, 2016, pp. 3124--3134. [Online].
  Available: \url{http://aclweb.org/anthology/C16-1294}
\BIBentrySTDinterwordspacing

\bibitem{Graham-etal:2014}
\BIBentryALTinterwordspacing
Y.~Graham, T.~Baldwin, A.~Moffat, and J.~Zobel, ``Is machine translation
  getting better over time?'' in \emph{Proceedings of the 14th Conference of
  the European Chapter of the Association for Computational Linguistics}.\hskip
  1em plus 0.5em minus 0.4em\relax Gothenburg, Sweden: Association for
  Computational Linguistics, 2014, pp. 443--451. [Online]. Available:
  \url{http://www.aclweb.org/anthology/E14-1047}
\BIBentrySTDinterwordspacing

\bibitem{Graham-etal:2016-da}
Y.~Graham, A.~Baldwin, Timothy~Moffat, and J.~Zobel, ``Can machine translation
  systems be evaluated by the crowd alone,'' \emph{Natural Language
  Engineering}, vol.~23, no.~1, pp. 3--30, 2016.

\bibitem{Graham-etal:2017}
\BIBentryALTinterwordspacing
Y.~Graham, Q.~Ma, T.~Baldwin, Q.~Liu, C.~Parra, and C.~Scarton, ``Improving
  evaluation of document-level machine translation quality estimation,'' in
  \emph{Proceedings of the 15th Conference of the European Chapter of the
  Association for Computational Linguistics: Volume 2, Short Papers}.\hskip 1em
  plus 0.5em minus 0.4em\relax Valencia, Spain: Association for Computational
  Linguistics, 2017, pp. 356--361. [Online]. Available:
  \url{http://www.aclweb.org/anthology/E/E17/E17-2057.pdf}
\BIBentrySTDinterwordspacing

\bibitem{Bentivogli-etal:2018}
\BIBentryALTinterwordspacing
L.~Bentivogli, M.~Cettolo, M.~Federico, and C.~Federmann, ``Machine translation
  human evaluation: an investigation of evaluation based on post-editing and
  its relation with direct assessment,'' in \emph{Proceedings of the 15th
  International Workshop on Spoken Language Translation}, Bruges, Belgium,
  2018, pp. 62--69. [Online]. Available:
  \url{https://workshop2018.iwslt.org/downloads/Proceedings_IWSLT_2018.pdf}
\BIBentrySTDinterwordspacing

\bibitem{Bojar2016}
\BIBentryALTinterwordspacing
O.~Bojar, R.~Chatterjee, C.~Federmann, Y.~Graham, B.~Haddow, M.~Huck, A.~J.
  Yepes, P.~Koehn, V.~Logacheva, C.~Monz, M.~Negri, A.~Neveol, M.~Neves,
  M.~Popel, M.~Post, R.~Rubino, C.~Scarton, L.~Specia, M.~Turchi, K.~Verspoor,
  and M.~Zampieri, ``{Findings of the 2016 Conference on Statistical Machine
  Translation},'' in \emph{Proceedings of the First Conference on Statistical
  Machine Translation}.\hskip 1em plus 0.5em minus 0.4em\relax Berlin, Germany:
  Association for Computational Linguistics, 2016, pp. 131--198. [Online].
  Available: \url{http://aclweb.org/anthology/W16-2301}
\BIBentrySTDinterwordspacing

\bibitem{graham-etal:2015}
\BIBentryALTinterwordspacing
Y.~Graham, T.~Baldwin, and N.~Mathur, ``{Accurate Evaluation of Segment-level
  Machine Translation Metrics},'' in \emph{Proceedings of the Human Language
  Technologies: The 2015 Annual Conference of the North American Chapter of the
  ACL}.\hskip 1em plus 0.5em minus 0.4em\relax Dever, CO: Association for
  Computational Linguistics, 2015, pp. 1183--119. [Online]. Available:
  \url{https://www.aclweb.org/anthology/N15-1124}
\BIBentrySTDinterwordspacing

\bibitem{Aziz-etal:2012}
\BIBentryALTinterwordspacing
W.~Aziz, S.~C.~M. Sousa, and L.~Specia, ``{PET: a tool for post-editing and
  assessing machine translation},'' in \emph{Proceedings of the 8th
  International Conference on Language Resources and Evaluation}, Istanbul,
  Turkey, 2012, pp. 3982--3987. [Online]. Available:
  \url{http://www.lrec-conf.org/proceedings/lrec2012/pdf/985_Paper.pdf}
\BIBentrySTDinterwordspacing

\bibitem{Papineni2002}
\BIBentryALTinterwordspacing
K.~Papineni, S.~Roukos, T.~Ward, and W.~jing Zhu, ``{BLEU: a Method for
  Automatic Evaluation of Machine Translation},'' in \emph{Proceedings of the
  40th Annual Meeting of the Association for Computational Linguistics}.\hskip
  1em plus 0.5em minus 0.4em\relax Philadelphia, PA: Association for
  Computational Linguistics, 2002, pp. 311--318. [Online]. Available:
  \url{https://www.aclweb.org/anthology/P02-1040.pdf}
\BIBentrySTDinterwordspacing

\bibitem{Banerjee2005}
\BIBentryALTinterwordspacing
S.~Banerjee and A.~Lavie, ``{METEOR: An Automatic Metric for MT Evaluation with
  Improved Correlation with Human Judgments},'' in \emph{Proceedings of the ACL
  2005 Workshop on Intrinsic and Extrinsic Evaluation Measures for MT and/or
  Summarization}.\hskip 1em plus 0.5em minus 0.4em\relax Ann Harbor, MI:
  Association for Computation Linguistics, 2005, pp. 65--72. [Online].
  Available: \url{http://aclweb.org/anthology/W05-0909}
\BIBentrySTDinterwordspacing

\bibitem{PBML_Asiya:2010}
J.~Gim\'{e}nez and L.~M\`{a}rquez, ``{Asiya: An Open Toolkit for Automatic
  Machine Translation (Meta-)Evaluation},'' \emph{The Prague Bulletin of
  Mathematical Linguistics}, no.~94, pp. 77--86, 2010.

\bibitem{callisonburch-EtAl:2012:WMT}
\BIBentryALTinterwordspacing
C.~Callison-Burch, P.~Koehn, C.~Monz, M.~Post, R.~Soricut, and L.~Specia,
  ``Findings of the {2012} {Workshop} on {Statistical} {Machine}
  {Translation},'' in \emph{Proceedings of the Seventh Workshop on Statistical
  Machine Translation}.\hskip 1em plus 0.5em minus 0.4em\relax Montr{\'e}al,
  Canada: Association for Computational Linguistics, June 2012, pp. 10--51.
  [Online]. Available: \url{http://www.aclweb.org/anthology/W12-3102}
\BIBentrySTDinterwordspacing

\bibitem{graham:2015}
\BIBentryALTinterwordspacing
Y.~Graham, ``{Improving Evaluation of Machine Translation Quality
  Estimation},'' in \emph{Proceedings of the 53rd Annual Meeting of the
  Association for Computational Linguistics and the 7th International Joint
  Conference on Natural Language Processing}.\hskip 1em plus 0.5em minus
  0.4em\relax Beijing, China: Association for Computational Linguistics, 2015,
  pp. 1804--1813. [Online]. Available:
  \url{http://www.aclweb.org/anthology/P15-1174}
\BIBentrySTDinterwordspacing

\bibitem{Koponen2012a}
M.~Koponen, W.~Aziz, L.~Ramos, and L.~Specia, ``Post-editing time as a measure
  of cognitive effort,'' in \emph{Proceedings of the AMTA 2012 Workshop on
  Post-Editing Technology and Practice}, San Diego, CA, 2012, pp. 11--20.

\bibitem{Plitt2010}
\BIBentryALTinterwordspacing
M.~Plitt and F.~Masselot, ``{A Productivity Test of Statistical Machine
  Translation Post-Editing in a Typical Localisation Context},'' \emph{The
  Prague Bulletin of Mathematical Linguistics}, vol.~93, pp. 7--16, 2010.
  [Online]. Available:
  \url{https://ufal.mff.cuni.cz/pbml/93/art-plitt-masselot.pdf}
\BIBentrySTDinterwordspacing

\bibitem{likert1932technique}
R.~Likert, ``A technique for the measurement of attitudes.'' \emph{Archives of
  psychology}, vol. 140, pp. 1--55, 1932.

\bibitem{Specia2011}
L.~Specia, N.~Hajlaoui, C.~Hallet, and W.~Aziz, ``Predicting machine
  translation adequacy,'' in \emph{Proceedings of the Machine Translation
  Summit XIII}, Xiamen, China, September 2011, pp. 19--23.

\bibitem{Bojar2014}
\BIBentryALTinterwordspacing
O.~Bojar, C.~Buck, C.~Federman, B.~Haddow, P.~Koehn, J.~Leveling, C.~Monz,
  P.~Pecina, M.~Post, H.~Saint-Amand, R.~Soricut, L.~Specia, and A.~Tamchyna,
  ``{Findings of the 2014 Workshop on Statistical Machine Translation},'' in
  \emph{Proceedings of the Ninth Workshop on Statistical Machine
  Translation}.\hskip 1em plus 0.5em minus 0.4em\relax Baltimore, MD:
  Association for Computational Linguistics, June 2014, pp. 12--58. [Online].
  Available: \url{http://www.aclweb.org/anthology/W/W14/W14-3302}
\BIBentrySTDinterwordspacing

\bibitem{OBrien2011}
S.~O'Brien, ``Towards predicting post-editing productivity,'' \emph{Machine
  Translation}, vol.~25, pp. 197--215, 2011.

\bibitem{Turian2003}
J.~P. Turian, L.~Shen, and I.~D. Melamed, ``{Evaluation of Machine Translation
  and its Evaluation},'' in \emph{Proceedings of the Machine Translation Summit
  IX}, New Orleans, LA, 2003, pp. 386--393.

\bibitem{Lacruz2014}
I.~Lacruz, M.~Denkowski, and A.~Lavie, ``{Cognitive Demand and Cognitive Effort
  in Post-Editing},'' in \emph{Proceedings of the Third Workshop on
  Post-Editing Technology and Practice}, Vancouver, Canada, 2014, pp. 73--84.

\bibitem{Bojar2013}
\BIBentryALTinterwordspacing
O.~Bojar, C.~Buck, C.~Callison-Burch, C.~Federmann, B.~Haddow, P.~Koehn,
  C.~Monz, M.~Post, R.~Soricut, and L.~Specia, ``Findings of the 2013 {Workshop
  on Statistical Machine Translation},'' in \emph{Proceedings of the Eighth
  Workshop on Statistical Machine Translation}.\hskip 1em plus 0.5em minus
  0.4em\relax Sofia, Bulgaria: Association for Computational Linguistics,
  August 2013, pp. 1--44. [Online]. Available:
  \url{http://www.aclweb.org/anthology/W13-2201}
\BIBentrySTDinterwordspacing

\bibitem{Bojar2015}
\BIBentryALTinterwordspacing
O.~Bojar, R.~Chatterjee, C.~Federmann, B.~Haddow, M.~Huck, C.~Hokamp, P.~Koehn,
  V.~Logacheva, C.~Monz, M.~Negri, M.~Post, C.~Scarton, L.~Specia, and
  M.~Turchi, ``{Findings of the 2015 Workshop on Statistical Machine
  Translation},'' in \emph{Proceedings of the Tenth Workshop on Statistical
  Machine Translation}.\hskip 1em plus 0.5em minus 0.4em\relax Lisbon,
  Portugal: Association for Computational Linguistics, 2015, pp. 1--46.
  [Online]. Available: \url{http://aclweb.org/anthology/W15-3001.pdf}
\BIBentrySTDinterwordspacing

\bibitem{Bojar2017}
\BIBentryALTinterwordspacing
O.~Bojar, R.~Chatterjee, C.~Federmann, Y.~Graham, B.~Haddow, S.~Huang, M.~Huck,
  P.~Koehn, Q.~Liu, V.~Logacheva, C.~Monz, M.~Negri, M.~Post, R.~Rubino,
  L.~Specia, and M.~Turchi, ``{Findings of the 2017 Conference on Machine
  Translation (WMT17)},'' in \emph{Proceedings of the Second Conference on
  Machine Translation, Volume 2: Shared Task Papers}.\hskip 1em plus 0.5em
  minus 0.4em\relax Copenhagen, Denmark: Association for Computational
  Linguistics, September 2017, pp. 169--214. [Online]. Available:
  \url{http://www.aclweb.org/anthology/W17-4717}
\BIBentrySTDinterwordspacing

\bibitem{Specia2018}
\BIBentryALTinterwordspacing
L.~Specia, F.~Blain, V.~Logacheva, R.~Astudillo, and A.~F.~T. Martins,
  ``Findings of the {WMT} 2018 shared task on quality estimation,'' in
  \emph{Proceedings of the Third Conference on Machine Translation, Volume 2:
  Shared Task Papers}.\hskip 1em plus 0.5em minus 0.4em\relax Belgium,
  Brussels: Association for Computational Linguistics, October 2018, pp.
  702--722. [Online]. Available: \url{http://www.aclweb.org/anthology/W18-6452}
\BIBentrySTDinterwordspacing

\end{thebibliography}
\bibliographystyle{IEEEtran}

\end{document}